\documentclass[a4paper, 11pt]{article}

\usepackage[english]{babel}
\usepackage[utf8]{inputenc}
\usepackage[T1]{fontenc}

\usepackage[a4paper,top=3cm,bottom=2cm,left=3cm,right=3cm,marginparwidth=1.75cm]{geometry}

\usepackage{amsmath}
\usepackage{graphicx}
\usepackage{booktabs}
\usepackage{caption}
\usepackage{subcaption}
\usepackage{float}
\usepackage{hyperref}
\usepackage{mathtools}
\usepackage[table]{xcolor}
\usepackage{multirow}
\usepackage{authblk}

\usepackage[backend=biber,style=numeric,sorting=none]{biblatex}
\hypersetup{
    colorlinks=true,
    allcolors=blue,
}
\definecolor{palegreen}{rgb}{0.92, 0.98, 0.92}

\title{\textbf{How much speech data is necessary for ASR in African languages? An evaluation of data scaling in Kinyarwanda and Kikuyu}}

\vspace{0.5em}

\author{Benjamin Akera}
\author{Evelyn Nafula}
\author{Patrick Walukagga}
\author{Gilbert Yiga}
\author{John Quinn}
\author{Ernest Mwebaze}

\affil{\textbf{Sunbird AI}}

\date{\today}

\begin{document}

\maketitle

\begin{abstract}
The development of Automatic Speech Recognition (ASR) systems for low-resource African languages remains challenging due to limited transcribed speech data. While recent advances in large multilingual models like OpenAI's Whisper offer promising pathways for low-resource ASR development, critical questions persist regarding practical deployment requirements. This paper addresses two fundamental concerns for practitioners: determining the minimum data volumes needed for viable performance and characterizing the primary failure modes that emerge in production systems. We evaluate Whisper's performance through comprehensive experiments on two Bantu languages: systematic data scaling analysis on Kinyarwanda using training sets from 1 to 1,400 hours, and detailed error characterization on Kikuyu using 270 hours of training data. Our scaling experiments demonstrate that practical ASR performance (WER < 13\%) becomes achievable with as little as 50 hours of training data, with substantial improvements continuing through 200 hours (WER < 10\%). Complementing these volume-focused findings, our error analysis reveals that data quality issues, particularly noisy ground truth transcriptions, account for 38.6\% of high-error cases, indicating that careful data curation is as critical as data volume for robust system performance. These results provide actionable benchmarks and deployment guidance for teams developing ASR systems across similar low-resource language contexts. We release accompanying and models see\footnote{\url{https://github.com/SunbirdAI/kinyarwanda-whisper-eval}}

\end{abstract}

\section{Introduction}

Automatic Speech Recognition technology has fundamentally transformed human-computer interaction, enabling voice assistants, transcription services, and accessibility tools that are now integral to modern digital infrastructure. Despite these advances, the benefits remain concentrated in a handful of high-resource languages, creating a pronounced digital divide that particularly affects African language communities. With over 2,000 distinct language varieties across the continent \cite{unesco2025promotion}, the scarcity of large transcribed speech corpora has historically prevented the development of effective ASR systems, limiting access to voice-enabled technologies for hundreds of millions of speakers.

The emergence of large-scale multilingual models has begun to reshape this landscape. OpenAI's Whisper \cite{radford2023robust}, trained on 680,000 hours of multilingual data, demonstrates remarkable zero-shot capabilities across diverse languages and can be fine-tuned with substantially less data than traditional approaches required. Similarly, advances in self-supervised learning \cite{jaiswal2020survey} have shown that models can leverage vast amounts of unlabeled audio to reduce supervised data requirements. These developments suggest that the data barriers to ASR development may be lowering significantly.

However, translating these technological advances into practical deployments requires addressing questions that remain largely unanswered in the literature. For language communities and researchers working with limited resources, fundamental uncertainties persist: What constitutes sufficient training data for achieving viable ASR performance in real-world applications? Beyond data volume, what are the primary sources of system failures, and how can practitioners anticipate and mitigate these issues during development?

This paper provides empirical answers to these deployment-critical questions through comprehensive evaluation of Whisper's performance on two representative Bantu languages: Kinyarwanda and Kikuyu. We contribute complementary analyses that together paint a complete picture of both opportunities and challenges in low-resource ASR development. Our systematic data scaling experiments on Kinyarwanda establish clear quantitative relationships between training data volume and recognition performance across eight different dataset sizes. Concurrently, our detailed error analysis of a Kikuyu system trained on 270 hours reveals the qualitative factors that drive system failures beyond simple data scarcity. These findings demonstrate that while practical ASR systems can indeed be developed with moderate data investments of 50-200 hours, success depends critically on maintaining data quality standards throughout the scaling process. Our results provide concrete benchmarks and actionable insights for teams developing speech technologies across African languages and similar low-resource contexts.

\section{Related Work}

The evolution of ASR for African languages reflects broader trends in speech recognition research, with each technological generation bringing both new capabilities and persistent challenges. Early efforts in the 2000s and 2010s relied on traditional statistical approaches, employing Gaussian Mixture Models and Hidden Markov Models that required extensive linguistic expertise for feature engineering and pronunciation modeling \cite{menon2018fast}, \cite{menon2018automatic}. These systems typically demanded hundreds of hours of carefully transcribed speech to achieve even modest performance levels, making them impractical for most African languages where such resources were unavailable.
The advent of deep learning fundamentally transformed ASR capabilities through end-to-end neural architectures. Attention-based models \cite{chorowski2015attention} eliminated many hand-crafted components while achieving superior performance, though often at the cost of requiring even larger training datasets. For low-resource languages, this created a paradox: more powerful models became available precisely when data requirements were increasing beyond the reach of most language communities.

Community-driven initiatives emerged to address these resource challenges systematically. The Masakhane research collective \cite{orife2020masakhane} has been particularly influential in coordinating dataset creation and model development across African languages, demonstrating how collaborative approaches can overcome individual resource constraints. These efforts have been complemented by advances in self-supervised learning, where models like wav2vec 2.0 \cite{baevski2020wav2vec} learn robust speech representations from unlabeled audio before fine-tuning on transcribed data. Such approaches have shown remarkable effectiveness in reducing supervised data requirements \cite{ogunremi2023multilingual}, \cite{alabi2024afrihubert}. The current generation of large multilingual pre-trained models represents another paradigm shift toward more practical low-resource ASR. Google's Universal Speech Model \cite{zhang2023google} demonstrates that massive-scale multilingual pre-training can enable effective cross-lingual transfer, allowing models trained primarily on high-resource languages to generalize to related low-resource varieties. OpenAI's Whisper \cite{radford2023robust} has gained particular attention for combining multilingual capabilities with practical features like automatic punctuation and capitalization, eliminating post-processing steps that traditionally required additional engineering effort. While these advances are promising, most existing studies evaluate performance through standard academic benchmarks that may not reflect deployment realities. Critical questions about practical data requirements, failure mode analysis, and error characterization remain underexplored for African languages specifically \cite{alabi2024afrihubert}. Our work addresses these gaps by providing focused empirical evaluation that bridges academic capabilities with practical deployment needs, offering concrete guidance for teams building real-world systems.

\section{Methodology}

% \subsection{Model Architecture and Training}

All experiments utilized the Whisper large-v3 model, a 1.55 billion parameter encoder-decoder transformer architecture. We employed a standardized fine-tuning pipeline developed through the Sunbird African Language Technology (SALT) initiative \cite{sunbird_parallel_2023} \cite{akera2022machine}, which has been optimized across multiple East African language projects.
The training configuration used a learning rate of $1 \times 10^{-5}$, batch size of 32, and early stopping with a patience of 4,000 steps on validation loss. To enhance model robustness, we applied several data augmentation techniques: random noise injection, speed perturbation (0.9-1.1x), and downsampling of 5\% of training samples to 8 kHz to simulate telephone-quality speech. All models were trained on H100 or A100 GPUs.

\subsection{Datasets and Experimental Design}

\begin{itemize}
    \item \textbf{Kinyarwanda Data Scaling Study:} We utilized the Digital Umuganda competition dataset, containing approximately 263,000 audio recordings across five domains: Health, Government, Financial Services, Education, and Agriculture. After removing 7,000 mislabeled examples, the cleaned dataset comprised approximately 1,400 hours of single-speaker utterances with minimal background noise.

    For the scaling analysis, we created training subsets of 1, 50, 100, 150, 200, 500, 1,000 hours, and the full dataset. Each subset was used to fine-tune a separate Whisper model, with performance evaluated on a consistent held-out test set of 1,000 samples.

    \item \textbf{Kikuyu Error Analysis Study:} We fine-tuned Whisper on approximately 270 hours of curated Kikuyu speech data, then performed detailed performance analysis on 6,910 evaluation samples. We categorized results into four performance bands based on WER thresholds and applied systematic heuristic tagging to identify common error patterns in high-WER samples (WER $\geq$ 40\%).

    Our error analysis methodology employed three binary tags: (1) spurious repetition, when repeated word density exceeded 10\%; (2) very long text, when either reference or prediction exceeded 80 tokens; and (3) noisy ground truth, indicated by explicit noise markers, high non-letter character density (>15\%), or very short references with high WER.

\end{itemize}

\section{Results}

\subsection{Data Scaling Analysis for Kinyarwanda}

Table~\ref{tab:kinyarwanda_scaling} presents the comprehensive results of our data scaling experiments. The relationship between training data volume and performance follows a clear logarithmic improvement pattern, with the most substantial gains occurring in the first 200 hours of data.

\begin{table}
\centering
\caption{Kinyarwanda ASR performance scaling with training data volume}
\label{tab:kinyarwanda_scaling}
\small
\renewcommand{\arraystretch}{1.2}
\rowcolors{2}{white}{palegreen}
\begin{tabular}{@{}lrrrrrc@{}}
\toprule
\textbf{Data Volume} & \textbf{Training} & \textbf{WER} & \textbf{CER} & \textbf{Train} & \textbf{Eval} & \textbf{GPU} \\
& \textbf{Steps} & \textbf{(\%)} & \textbf{(\%)} & \textbf{Time (h)} & \textbf{Time (h)} & \textbf{Hardware} \\
\midrule
Baseline (0h) & 0 & 44.41 & 25.39 & -- & 4.4 & -- \\
1 hour & 4,400 & 47.63 & 16.97 & 3.9 & 4.4 & H100/A100 \\
50 hours & 5,600 & 12.51 & 3.31 & 4.5 & 3.8 & A100 \\
100 hours & 6,800 & 10.90 & 2.84 & 8.9 & 2.4 & A100 \\
150 hours & 8,000 & 10.21 & 2.64 & 10.6 & 2.4 & A100 \\
200 hours & 8,400 & 9.82 & 2.56 & 6.7 & 2.4 & H100 \\
500 hours & 15,200 & 8.24 & 2.15 & 11.9 & 2.4 & H100 \\
1,000 hours & 21,200 & 7.65 & 1.98 & 18.3 & 2.4 & H100 \\
Full dataset & 25,000 & 7.14 & 1.88 & 20.9 & 2.3 & H100 \\
\bottomrule
\end{tabular}
\end{table}

The most significant finding is that practical ASR performance becomes achievable with modest data volumes. The 50-hour model achieves 12.51\% WER, representing a 75\% improvement over the 1-hour baseline and reaching a level suitable for semi-automated transcription applications. Performance crosses the 10\% WER threshold at 200 hours (9.82\%), while continuing to improve even with the full 1,400-hour dataset (7.14\% WER).
Character Error Rate follows a similar pattern, improving from 16.97\% at 1 hour to 3.31\% at 50 hours, and reaching 1.88\% with the complete dataset. The consistent improvement across both metrics suggests that additional high-quality data continues to provide benefits even at large scales.

\begin{figure}
    \centering
    \includegraphics[width=0.8\linewidth]{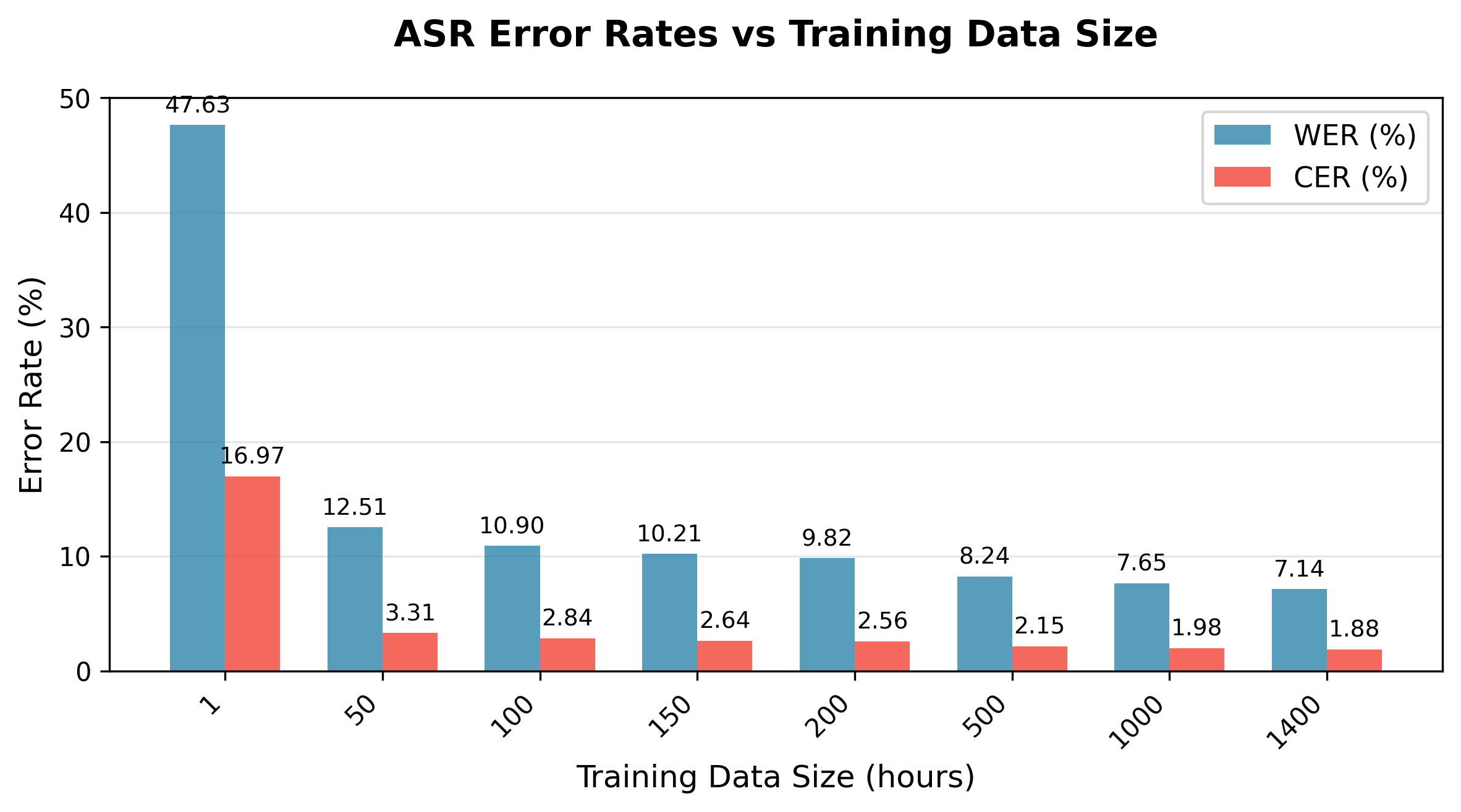}
    \caption{Word Error Rate and Character Error Rate by training data volume for Kinyarwanda. Both metrics show consistent improvement with increased data, with the most substantial gains in the first 200 hours.}
    \label{fig:kinyarwanda_error_rates}
\end{figure}

Figure~\ref{fig:kinyarwanda_error_rates} visualizes the relationship between training data volume and both WER and CER. The logarithmic improvement pattern is clearly evident, with the steepest gains occurring between 1 and 200 hours of training data.

\begin{figure}[t]
    \centering
    \includegraphics[width=\textwidth]{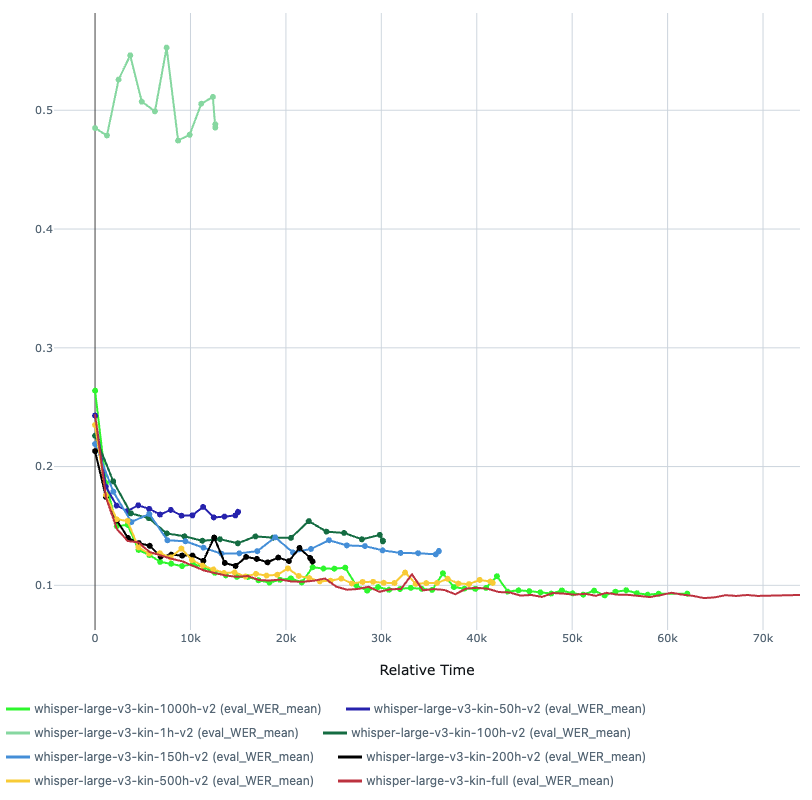}
    \caption{Validation Word Error Rate evolution during training for different Kinyarwanda dataset sizes. Training time is shown in seconds on a single H100 GPU, with the longest run requiring 20.2 hours.}
    \label{fig:wer_evolution}
\end{figure}

The training dynamics across different data scales are illustrated in Figure~\ref{fig:wer_evolution}, which shows validation WER progression during fine-tuning. Notably, models trained on larger datasets not only achieve lower final error rates but also demonstrate more stable convergence patterns.

\subsection{Performance Distribution Analysis for Kikuyu}

Our Kikuyu model, trained on 270 hours, achieved a median WER of 26.3\% and mean WER of 30.3\% across 6,910 evaluation samples. The gap between median and mean indicates a substantial tail of high-error samples driving overall performance statistics.

Table~\ref{tab:kikuyu_performance} presents the detailed performance breakdown across four quality bands. The distribution reveals a bimodal pattern: while 22.8\% of samples achieve excellent to good performance (WER < 10\%), the majority fall into an "acceptable" range with moderate errors.

\begin{table}
\centering
\caption{Kikuyu ASR performance distribution by quality bands}
\label{tab:kikuyu_performance}
\small
\renewcommand{\arraystretch}{1.2}
\rowcolors{2}{white}{palegreen}
\begin{tabular}{@{}lrrrrrr@{}}
\toprule
\textbf{Quality Band} & \textbf{Count} & \textbf{Share} & \textbf{Mean WER} & \textbf{Median WER} & \textbf{Min WER} & \textbf{Max WER} \\
& & \textbf{(\%)} & \textbf{(\%)} & \textbf{(\%)} & \textbf{(\%)} & \textbf{(\%)} \\
\midrule
Excellent & 1,231 & 17.8 & 0.08 & 0.00 & 0.00 & 5.0 \\
Good & 343 & 5.0 & 8.4 & 8.3 & 5.3 & 10.0 \\
Acceptable & 4,206 & 60.9 & 29.4 & 28.6 & 10.3 & 50.0 \\
Poor & 1,130 & 16.4 & 73.1 & 66.7 & 50.8 & 573.3 \\
\bottomrule
\end{tabular}
\end{table}

\begin{figure}
    \centering
    \includegraphics[width=0.9\linewidth]{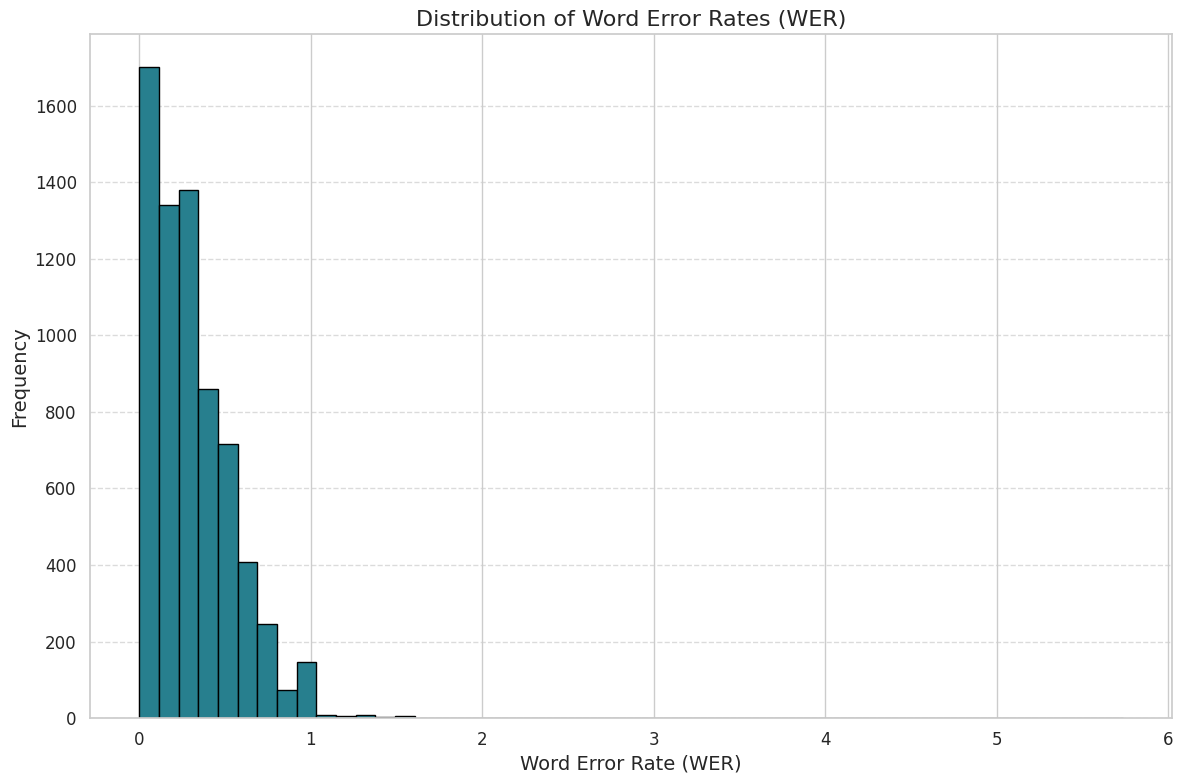}
    \caption{Distribution of Word Error Rates across 6,910 Kikuyu evaluation samples. The histogram shows high concentration in the 0-50\% range with a long tail of higher-error cases.}
    \label{fig:kikuyu_wer_dist}
\end{figure}

The distribution of error rates across the evaluation set, shown in Figure~\ref{fig:kikuyu_wer_dist}, reveals the concentration of performance in different ranges. While the majority of samples cluster in the 0-50\% WER range, a notable tail extends to much higher error rates.

\begin{figure}
    \centering
    \includegraphics[width=0.7\linewidth]{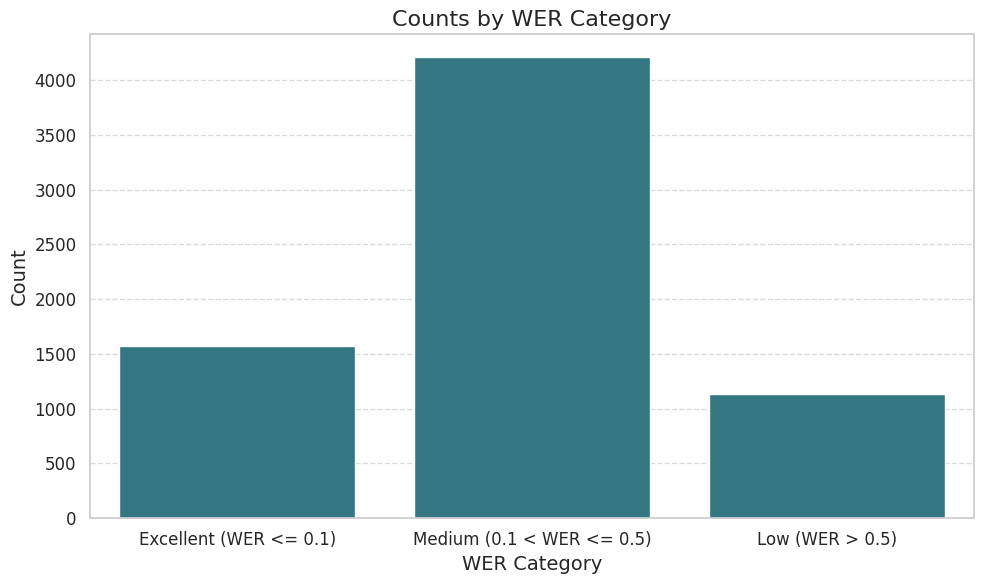}
    \caption{Word Error Rate distribution by performance category for Kikuyu evaluation data. Box plots show median, quartiles, and outliers for each quality band.}
    \label{fig:kikuyu_wer_category}
\end{figure}

This distribution pattern becomes clearer when viewed by performance categories, as illustrated in Figure~\ref{fig:kikuyu_wer_category}. The categorical breakdown demonstrates how the tail of poor performance substantially impacts overall statistics.

The "Low" category, representing 16.4\% of samples, exhibits extremely high error rates with a mean WER of 73.1\%. Some samples in this category show WER exceeding 100\%, indicating cases where the model generates substantially more text than the reference, often due to repetition artifacts.

\subsection{Error Mode Analysis}

To understand the drivers of poor performance, we analyzed all 2,187 samples with WER $\geq$ 40\%. Our heuristic tagging revealed systematic patterns in failure cases, as shown in Table~\ref{tab:error_modes}.

\begin{table}
\centering
\caption{Error mode distribution in high-WER Kikuyu samples (WER $\geq$ 40\%)}
\label{tab:error_modes}
\small
\renewcommand{\arraystretch}{1.2}
\begin{tabular}{@{}lrr@{}}
\toprule
\textbf{Error Mode} & \textbf{Count} & \textbf{Percentage} \\
\midrule
Noisy/unclear ground truth & 844 & 38.6\% \\
Very long text (>80 tokens) & 114 & 5.2\% \\
Spurious repetition & 96 & 4.4\% \\
Multiple factors & 78 & 3.6\% \\
Other/unclear & 1,055 & 48.2\% \\
\midrule
\textbf{Total analyzed} & \textbf{2,187} & \textbf{100.0\%} \\
\bottomrule
\end{tabular}
\end{table}

The most significant finding is that noisy or unclear ground truth accounts for 38.6\% of high-error cases. These samples typically contained explicit noise markers (e.g., "[noise]", "(inaudible)"), unusually high proportions of non-alphabetic characters, or extremely short references where single token errors result in disproportionately high WER values.
Long utterances (>80 tokens) contribute to 5.2\% of errors, likely due to alignment drift or attention mechanism limitations over extended sequences. Model-induced repetition loops account for 4.4\% of failures, representing a known pathology in autoregressive generation that could potentially be mitigated through improved decoding strategies.

\section{Discussion}

Our dual-language study provides complementary insights into both the opportunities and challenges of developing ASR for low-resource African languages using large pre-trained models.
The Kinyarwanda scaling results are encouraging for practitioners: viable ASR systems can be developed with data volumes that are achievable for many language communities. The 50-hour threshold for practical performance (12.51\% WER) represents a significant reduction from the hundreds or thousands of hours traditionally required for ASR development. This finding aligns with recent work on few-shot learning with large models, suggesting that Whisper's multilingual pre-training provides strong inductive biases for related languages.
However, our Kikuyu error analysis reveals that data quality becomes increasingly critical as systems scale. The finding that 38.6\% of high-error cases stem from noisy ground truth rather than model limitations highlights a fundamental challenge: as data collection scales, maintaining annotation quality becomes both more important and more difficult. This suggests that resource allocation should balance data volume with rigorous quality control processes.
The relatively low incidence of length-related (5.2\%) and repetition (4.4\%) errors indicates that these technical limitations, while notable, are secondary to data quality issues. This finding is practically significant because data quality problems can be addressed through improved collection and annotation protocols, while model architectural limitations would require more substantial interventions.
Several limitations constrain the generalizability of our findings. Our evaluation was conducted on relatively clean test sets that may not fully represent real-world acoustic conditions. The Kinyarwanda test data particularly lacks environmental noise and speaker diversity that would be encountered in deployment scenarios. Additionally, we did not explore transfer learning from related higher-resource languages, which could potentially improve sample efficiency for very small datasets.

\section{Conclusion}

This study demonstrates that effective ASR systems for low-resource African languages are achievable with moderate data investments when leveraging large pre-trained models like Whisper. Our Kinyarwanda scaling analysis establishes clear benchmarks: practical performance emerges around 50 hours of training data, with substantial improvements continuing through 200 hours and beyond. Equally important, our Kikuyu error analysis reveals that traditional metrics of "data volume" must be complemented by attention to data quality. The predominance of ground truth issues among failure cases suggests that successful ASR deployment requires not just sufficient data, but carefully curated data with consistent annotation standards.
For practitioners developing ASR systems for African languages, our findings suggest a two-phase approach: initial development can target the 50-200 hour range for viable performance, followed by systematic quality improvement of existing data rather than purely volume-driven expansion. This strategy optimizes resource allocation while avoiding the diminishing returns and error propagation that can result from poorly curated large datasets.

Future work should extend this analysis to more diverse acoustic conditions, explore cross-lingual transfer learning between related African languages, and develop automated quality assessment tools to support large-scale data curation efforts.

\section*{Acknowledgments}

The authors acknowledge the Digital Umuganda competition organizers for providing the Kinyarwanda dataset and the broader Masakhane community for fostering collaboration in African language technologies.

\section*{Code and Data Availability}

All code, models, and datasets used in this study are publicly available:

\begin{itemize}
\item Codebase: \url{https://github.com/SunbirdAI/kinyarwanda-whisper-eval}
\item Kinyarwanda model: Available on Hugging Face Hub
\item Kikuyu model: \url{https://huggingface.co/akera/whisper-large-v3-kik-full_v2}
\item Kikuyu evaluation data: \url{https://huggingface.co/datasets/evie-8/kikuyu-data}
\end{itemize}

\printbibliography

\end{document}